\documentclass[a4paper,10pt,conference]{ieeeconf}  

\IEEEoverridecommandlockouts                              
\usepackage{graphicx} 
\usepackage{amsmath} 
\usepackage{amssymb}  
\usepackage{algorithm}
\usepackage{algorithmic}
\usepackage{xcolor}
\usepackage{soul}
\usepackage{multirow}
\usepackage[font=footnotesize]{caption}
\usepackage[font=scriptsize]{subfig}

\let\labelindent\relax
\usepackage{enumitem}

\makeatletter
\let\NAT@parse\undefined
\makeatother
\usepackage{hyperref}
\usepackage{float}
\graphicspath{{figures/}}

\title{\LARGE \bf Evaluating Recipes Generated from Functional Object-Oriented Network}

\author{Md Sadman Sakib, Hailey Baez, David Paulius, and Yu Sun\\
\thanks{
Yu Sun, Md Sadman Sakib, and Hailey Baez are members of the Robot Perception and Action Lab (RPAL), which is part of the Department of Computer Science \& Engineering at the University of South Florida, Tampa, FL, USA. David is a postdoctoral researcher in the Human-centered Assistive Robotics group at the Technical University of Munich, Germany.
 \newline(Contact: \texttt{\{mdsadman,yusun\}@usf.edu)}}%
}

\begin{document}

\maketitle

\thispagestyle{empty}
\pagestyle{empty}

\begin{abstract}
The \textit{functional object-oriented network} (FOON) has been introduced as a knowledge representation, which takes the form of a graph, for symbolic task planning. To get  a sequential plan for a manipulation task, a robot can obtain a task tree through a knowledge retrieval process from the FOON. To evaluate the quality of an acquired task tree, we compare it with a conventional form of task knowledge, such as recipes or manuals. We first automatically convert task trees to recipes, and we then compare them with the human-created recipes in the Recipe1M+ dataset via a survey. Our preliminary study finds no significant difference between the recipes in Recipe1M+ and the recipes generated from FOON task trees in terms of correctness, completeness, and clarity.  
\end{abstract}

\section{Introduction}
In the development of intelligent and autonomous robots for different domains, a knowledge representation is a crucial component that serves as a bridge between human and robot understanding of tasks and action~\cite{paulius2019survey}.
Just as humans rely on manuals or recipes to provide knowledge on how to perform tasks, a robot can rely on a knowledge system to understand its domain.
To realize this goal, we introduced the functional object-oriented network (FOON) as a source of symbolic-level knowledge for service robots~\cite{paulius2016functional,paulius2018functional}.
FOON, which was inspired by the theory of affordance \cite{Gibson_1977} and prior work on joint object-action representation~\cite{Ren2013,SunRAS2013,Lin2015a}, takes the form of a bipartite graph to describe the relationship between objects and manipulation actions as nodes in the network.
With such a knowledge graph, a robot can acquire knowledge in the form of a task sequence with which it can determine how it should go about solving a manipulation problem, which would be reflected as a target node in a FOON.
Using a graph structure is beneficial for its interpretability as well as ease of use in knowledge retrieval.

Ideally, a FOON should provide concrete and correct knowledge to a robot for it to understand how to carry out a given task. The quality of this knowledge should be determined by how a robot executes the task. However, this is very challenging because of the variability in real world. A successful execution depends heavily on the robot's motor and sensor capacity, joint strength, degrees of freedom, etc.
Equally important, the knowledge should closely match to how a human may understand or approach the problem.
For instance, the cooking procedures described in FOON should be similar to those from widely available knowledge sources such as recipes and cookbooks, taking into account contextually relevant and accurate details for how objects should be manipulated or prepared.
With this vision in mind, our aim in this work is to evaluate the quality of our FOON graphs when compared to knowledge sources that are typically used by humans, such as cooking recipes (which we focus on in this work).
To accomplish this, we explore recipe generation, where a task sequence obtained from FOON can be translated into recipe-like instructions, which can then be compared to other text-based recipes.
Through a user study, we aim to determine whether there are any significant differences between knowledge obtained from FOON and conventional recipes for equivalent meals.
Recipes used in our evaluation were obtained from the Recipe1M+ dataset, which contains over 1 million recipes created by humans and that were shared on the internet~\cite{marin2019learning}.

\begin{figure}[t]
	\centering
	\includegraphics[trim={1.5cm 1.5cm 0.5cm 1cm},clip,width=8cm]{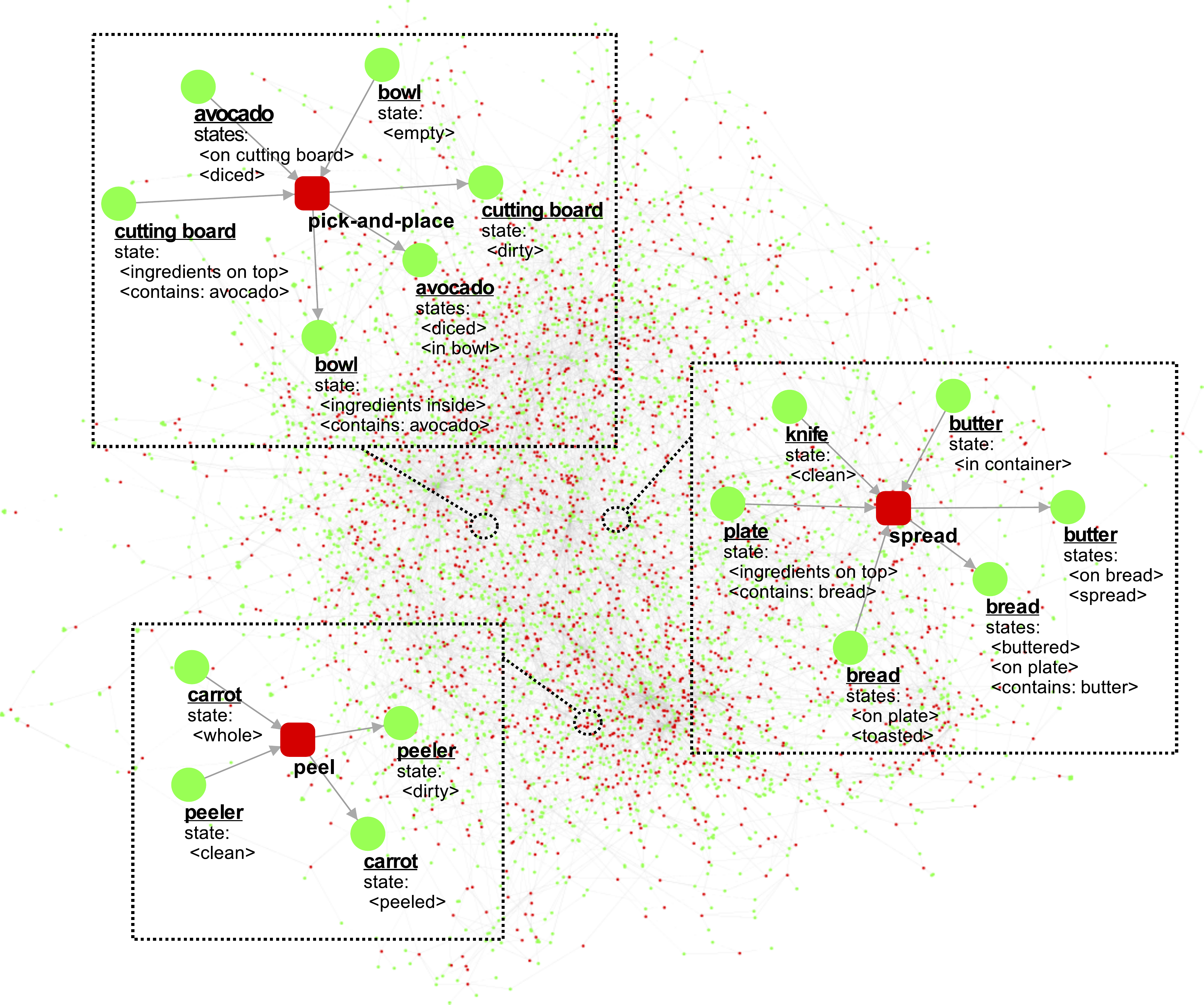}
	\caption{Illustration of a universal FOON made of 111 instructional videos. 
	This network, along with other subgraphs, are available on our website \cite{foonet}.}
	\label{fig:FOON}
\end{figure}

This paper is organized as follows: in Section \ref{sec:FOON}, we give a short overview of the FOON structure, our dataset, and the concept of task tree retrieval for task planning.
In Section \ref{sec:survey}, we introduce the user study that we conducted to evaluate the quality of our FOON graphs and discuss our observations from the study.
Finally, in Section \ref{sec:con}, we conclude our discussion and outline future work and directions.

\section{Functional Object-Oriented Network}
\label{sec:FOON}
FOON was originally introduced in \cite{paulius2016functional} as a graph-based knowledge representation that represents high-level concepts related to human manipulations for service robots.
To represent activities, a FOON contains two types of nodes, \textit{object nodes} and \textit{motion nodes}, making FOON a bipartite network.
Affordances are depicted with edges that connect objects to actions, which also indicate order of actions in the network.
The coupling of object and motion nodes to represent a single action is referred to as a \textit{functional unit}.

\begin{figure}[t]
	\centering
    \includegraphics[width=7cm]{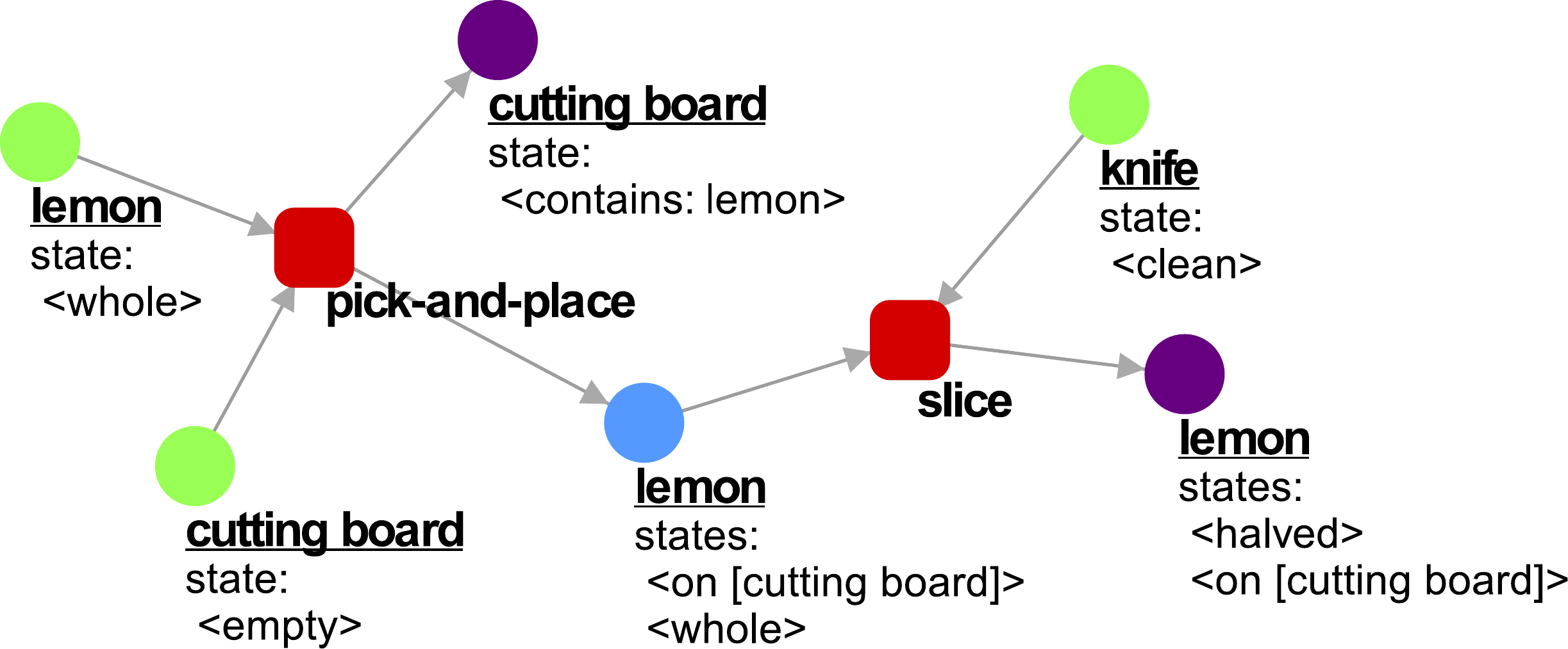}
	\caption{An example of two connected functional units (best viewed in colour), where object nodes are denoted by circles and motion nodes are denoted by squares. We show input-only nodes in green, output-only nodes in purple, and nodes that are both input and output to units in blue.
	}
	\label{fig:unit}
\end{figure}

\subsection{Defining and creating a FOON}
In FOON, we represent actions as a fundamental learning unit known as the functional unit, which contains object and motion nodes to vividly describe a single action within an activity.
In Figure \ref{fig:unit}, we show two functional units that describe the actions of placing a whole \textit{lemon} object onto a \textit{cutting board} and then slicing the \textit{lemon} with a \textit{knife} into halves.
Note that there exists multiple instances of object nodes due to the state transitions that occur as a result of an action; this is because a functional unit describes the state change of manipulated objects before and after execution.
Therefore, it is important to be able to detect states so that one can determine when an action has been completed~\cite{jelodar2019joint}.
{\it Input} object nodes describe the required state(s) of objects needed for the task, and {\it output} object nodes describe the outcome of the inputs for said task.
Some actions may not cause a change in all input objects' states, so there may be instances with fewer output nodes than inputs.


A FOON is created using annotations of video demonstrations, especially cooking videos from YouTube, and converting them into the FOON graph structure; in this annotation process, we note the actions, objects, and state changes (as functional units) that eventually result in a specific meal or product. 
A FOON that represents a single activity is referred to as a {\it subgraph}; a subgraph contains functional units in sequence to describe objects' states before and after each action occurs, the time-stamps when each action occurs in the demonstration, and what objects are being manipulated.
We cannot directly use existing text-based recipes because they do not contain the low-level instructions required for a robot to perform manipulations, especially since they rely on commonsense knowledge.
Presently, annotation is done manually by hand, but efforts have been made to investigate how it can be done as a semi-automatic process~\cite{jelodar2018long}.
Two or more subgraphs can be merged together to form a {\it universal} FOON, which is simply a FOON containing information from several sources of knowledge, and as such, a universal FOON could propose variations of recipes.
The merging procedure is simply a union operation done on all functional units from each subgraph we wish to combine; as a result, duplicate functional units are eliminated.

\subsection{Overview of Task Tree Retrieval}
A FOON can not only be used for representing knowledge, but a robot can use it for problem solving.
Given a goal, through the process of task tree retrieval, a robot can obtain a subgraph that contains functional units for actions it needs to execute to achieve it.
A subgraph that is obtained from knowledge retrieval is called a {\it task tree}.
A task tree differs from a regular subgraph, as it will not necessarily reflect a complete procedure from a single human demonstration.
Rather, it will leverage knowledge from multiple sources to produce a novel task sequence.
This search requires a list of items in its environment (i.e. the kitchen) to identify ideal functional units based on the availability of inputs to these units.
This algorithm is motivated by typical graph-based depth-first search (DFS) and breadth-first search (BFS): starting from the goal node, we search for candidate functional units in a depth-wise manner, while for each candidate unit, we search among its input nodes in a breadth-wise manner to determine whether or not they are in the kitchen.
Recently, we created an alternate retrieval algorithm that can take into account a FOON that has weights assigned to each functional unit to reflect their difficulty in robotic execution~\cite{paulius2021weighted}.

\section{Evaluation Approach}
\label{sec:survey}


In this section, first, we present a method to generate a recipe from a task tree. We then address the survey questions in detail. Finally, we analyze the survey responses with statistical tests and report our findings. 

\subsection{Recipe Generation from a Task Tree}
\begin{figure}[t]
	\centering
	\subfloat[Subgraph of preparing scrambled eggs]{{\includegraphics[width=6.5cm, height=8.9cm]{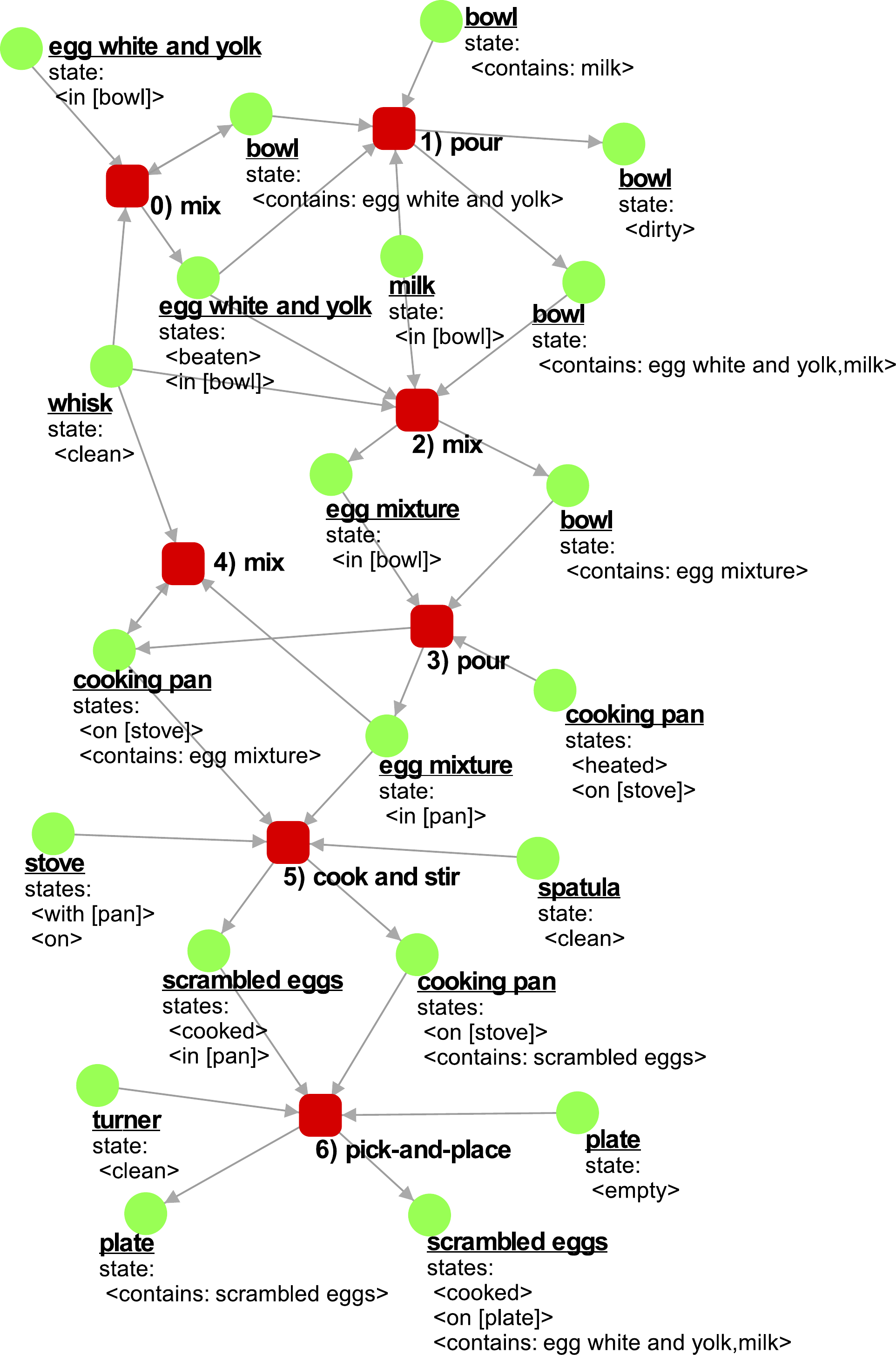}}
    \label{fig:gen-1}}    
	
	\subfloat[Generated recipe for preparing scrambled eggs]{{
		\noindent 
		\fbox{\parbox{.9\columnwidth}
		{\footnotesize
		   \textbf{Instructions:}
		       \begin{enumerate}
		            \item mix 4 egg white and yolk 
                    \item pour beaten egg white and yolk, 2 tsp milk to bowl 
                    \item mix beaten egg white and yolk, milk 
                    \item pour egg mixture from bowl to cooking pan 
                    \item mix egg mixture 
                    \item cook and stir egg mixture 
                    \item place cooked scrambled eggs from pan to plate 
		       \end{enumerate}
        }
	}}
	\label{fig:gen-2}}
	
	\subfloat[Recipe for preparing scrambled eggs from Recipe1M+]{{
		\noindent 
		\fbox{\parbox{.95\columnwidth}
		{\footnotesize
		   \textbf{Instructions:}
		       \begin{enumerate}
		             \item dice onion finely
                    \item dice tomato into small pieces
                    \item fry onion in oil until it brown
                    \item not all onions will brown however so be careful not to burn it
                    \item add ginger-garlic paste and tomato and fry for about 30 seconds
                    \item add eggs and salt 
                    \item turn down heat to low
                    \item scramble the eggs by stirring until desired consistency is achieved
                    \item enjoy while hot
		       \end{enumerate}
        }
	}}
	\label{fig:gen-3}}

	\caption{Example of: (a) a FOON subgraph, (b) its equivalent generated recipe, and (c) a similar example from Recipe1M+ for preparing eggs. Recipes such as (b) and (c) were provided to survey participants.}
	\label{fig:gen}
\end{figure}

The objective of recipe generation is to translate a subgraph or task tree into an instructional format for easy reading and verification, based on the following rules:

\begin{itemize}
    \item Each functional unit is translated into a descriptive sentence, where each generated sentence is in the form of \{\textless motion label\textgreater, \textless portion\textgreater, \textless object state\textgreater, \textless ingredient label\textgreater, \textless additional information\textgreater\}.
    
    \item For portion information, we rely on external sources, as a FOON does not regularly provide them. We only add the portion for the first instance of an object in a recipe.
    
    \item The additional information varies depending on the motion being applied. For actions that suggest a source and destination (such as pouring, adding or placing), we include source and target names based on the motion identifier flag, which is assigned to each object in a FOON, whereas utensil information is used in case of actions such as stirring or mixing.
    
    \item We merge consecutive sentences if possible (e.g., `pour milk', `pour butter', `pour olive oil' can be merged to `pour milk, butter and olive oil'). Moreover, we remove some sentences that are not instructive for humans (e.g., clean utensil or container becomes dirty after using it).
\end{itemize}
In Figure \ref{fig:gen}, the two recipes for preparing eggs have some different ingredients and cooking steps because they originate from different sources. Still, they can be called equivalent if they get similar ratings from the survey participants.


\subsection{Survey Objectives and Questions}
The goal of this survey is to evaluate the quality and completeness of the recipes generated from FOON. The idea is to compare FOON recipes with Recipe1M+ and measure how significant their differences are based on how they are interpreted by humans. To do this, we have created a pool of 111 FOON recipes and 111 equivalent recipes from Recipe1M+, which were selected based on an overlapping of main ingredients and sharing a similar recipe type. There were a total of nine participants who participated in this survey. Each participant was provided 10 recipes randomly selected from those 222 recipes without mentioning which recipe is coming from which dataset. 
We provide an overview of these survey questions in Figure~\ref{fig:questions}. 
At first, we asked a few questions regarding their cooking expertise (as questions 1-3). The remainder of the questions were designed to evaluate the correctness, completeness, and clarity of the recipe. In detail, participants had to answer questions 4-7 by rating them from 1 to 10. There was also an option to skip questions if the participant is not sure about how to answer the question. The expectation is that Recipe1M+ recipes and FOON-generated recipes will not differ greatly in terms of the ratings, which should be evident in the statistical tests such as Student's $t$-Test \cite{student08ttest}.
\newline

\begin{figure}[t]
    \centering
	\noindent \fbox{\parbox{0.95\columnwidth}
		{\footnotesize
		   \textbf{Baseline Questions:}
		   \begin{enumerate}
		       \item Please describe your proficiency level in cooking.
		       \begin{itemize}[label=$\Box$]
		           \item I have no experience in cooking
		           \item Beginner home cook
		           \item Intermediate home cook
		           \item Advanced home cook
		           \item I have received culinary training
		       \end{itemize}
		       \item If or when you cook, what type of recipes do you use?
		       \begin{itemize}[label=$\Box$]
		           \item I mostly use recipes that family or friends shared
		           \item I look for recipes online
		           \item I follow recipes from cookbooks
		           \item I only use ingredients I have available
		       \end{itemize}
		       \item Have you made a dish similar to this recipe? 
		       \begin{itemize}[label=$\Box$]
		           \item I have made this exact dish
		           \item Yes, but I left out some of the ingredients listed
		           \item Yes, but I added some ingredients not listed
		           \item No
		       \end{itemize}
		   \end{enumerate}
		   \textbf{Questions on Correctness of Recipe:}
		   \begin{enumerate}
		   \setcounter{enumi}{3}
		       \item Does this recipe seem correct to you?
		       \item Do the steps appear to be in the right order? 
		       \item Are all steps in the recipe correct? 
		   \end{enumerate}
		   \textbf{Questions on Completeness of Recipe:}
		   \begin{enumerate}
		   \setcounter{enumi}{6}
		       \item Does each recipe step give enough information in order to complete it?
		       \item Does the recipe skip steps that are obvious to you? 
		   \end{enumerate}
		   \textbf{Questions on Clarity of Recipe:}
		   \begin{enumerate}
		   \setcounter{enumi}{8}
		       \item If attempting to follow the given steps, how confident are you that you could make this dish?
		       \item Do the steps in this recipe appear to be clear and easy to follow?
		   \end{enumerate}
		   
	}}
	\caption{Overview of the survey questions used in our human study.}
	\label{fig:questions}
\end{figure}

\section{Results and Discussion}
\begin{figure}[t]
	\centering
	\includegraphics[width=\columnwidth]{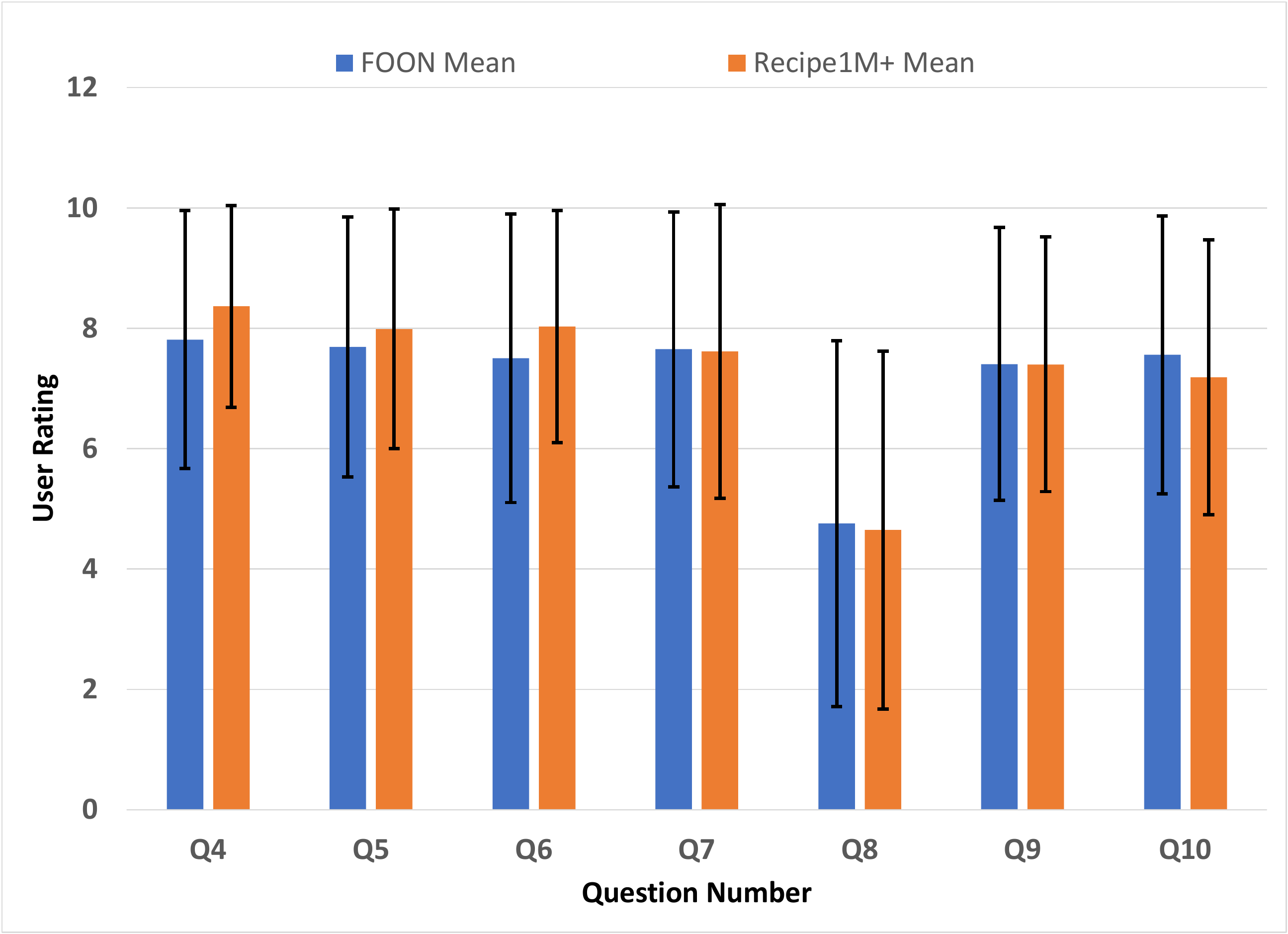}
	\caption{Comparison of weighted mean between the ratings for both FOON and Recipe1M+ recipes. Standard deviations are reported with error bars. 
    }
	\label{fig:mean_std_deviation}
\end{figure}

We computed the weighted mean $\Bar{x}$ and weighted standard deviation $\sigma$ on ratings for each of questions 4-7 separately. 
The weights are assigned based on the answers of questions 1-3, where participants with higher cooking proficiency or expertise are assigned more weight to their ratings compared to those having little to no cooking knowledge. $\Bar{x}$ and $\sigma$ are defined as follows:  
\begin{align*}
    & \Bar{x} = \frac{\sum_{i=1}^{n} w_i x_i}{\sum_{i=1}^{n} w_i}, 
    & \sigma = \frac{\sum_{i=1}^{n} w_i (x_i - \Bar{x})^2}{\frac{(n-1) \sum_{i=1}^{n} w_i}{n}}
\end{align*}
where $x_i$ is the $i$-th rating, $w_i$ is the weight applied on $x_i$ and $n$ is the number of ratings received for a question. $\Bar{x}$ and $\sigma$ values for Recipe1M+ and FOON-generated recipes are reported in Figure~\ref{fig:mean_std_deviation}. 
In terms of completeness and clarity, we observed that FOON-generated recipes have a slightly better average rating, whereas Recipe1M+ has the superior result for correctness. 
However, the mean value is not sufficient evidence to decide which dataset is better since a sample with a higher mean can have worse performance because of high standard deviation. A null hypothesis test \cite{fisher:1935} and equivalence test can help us to determine the statistical significance between two samples.
Our hypothesis, $H_o$ is that there is no significant difference between FOON-generated and Recipe1M+ recipes. We test it using a two-tailed independent $t$-Test and Two One-Sided Tests (TOST) with the values from Figure \ref{fig:mean_std_deviation}. According to null hypothesis,
\begin{align*}
    &\textbf{if } \text{$p$-value $> \alpha$: $H_o$ cannot be rejected} \\
    &\textbf{else: } \text{$H_o$ is rejected}
\end{align*}

Since all the $p$-values in Table \ref{table:TOST} are fairly greater than $\alpha$, we cannot reject $H_o$. Also, the equivalence bounds and confidence intervals (CI) in Table \ref{table:TOST} clearly shows the statistical equivalence between the two means. Based on the two tests combined, we can conclude that $H_o$ holds. The very high value in Q9 can be interpreted as the participants are equally confident about preparing a dish from FOON-generated and Recipe1M+ recipes.
This is promising, as this suggests that FOON subgraphs accurately depict cooking procedures.

\section{Conclusion and Future Work}
\label{sec:con}
In summary, we introduced a method to generate recipe-like instructions from FOON task trees. The primary purpose of the recipe generation process is to open up the possibility of evaluating FOON task trees by comparing them to textually rich data sources. We compared them with equivalent examples from the Recipe1M+ dataset to evaluate the quality of the generated recipes. Through a survey, we collected user ratings on recipes from both FOON and Recipe1M+, which were used to run Student's $t$-Test and equivalence test. Our investigation shows evidence that there is no significant difference between these two types of recipes. 

\begin{table}[t]
\centering
\caption{Results of null hypothesis ($H_o$) test and equivalence test using the mean and standard deviation from Figure~\ref{fig:mean_std_deviation} with significance level $\alpha=0.05$, standardised difference Cohen’s d = 0.3}
\begin{tabular}{|c|c|c|c|}
\hline
\multirow{2}{*}{\textbf{Question}} & \textbf{$H_o$ Test} & \multicolumn{2}{|c|}{\textbf{Equivalence Test}}\\ \cline{2-4}
     & p-value & Equivalence bounds & 90\% TOST CI\\ \hline
$Q4$ & $0.21$ & $[-0.58,0.58]$ & $[-1.25,0.14]$  \\ \hline
$Q5$ & $0.51$ & $[-0.62,0.62]$ & $[-1.04,0.44]$  \\ \hline
$Q6$ & $0.28$ & $[-0.65,0.65]$ & $[-1.31,0.26]$  \\ \hline
$Q7$ & $0.94$ & $[-0.71,0.71]$ & $[-0.81,0.89]$  \\ \hline
$Q8$ & $0.88$ & $[-0.90, 0.90]$ & $[-1.08,1.30]$  \\ \hline
$Q9$ & $0.99$ & $[-0.66,0.60]$ & $[-0.77,0.78]$  \\ \hline
$Q10$ & $0.45$ & $[-0.69,0.69]$ & $[-0.44,1.19]$  \\ \hline
\end{tabular}
\label{table:TOST}
\end{table}

In the future, we will further explore the recipe generation procedure to create more dynamic and instructive instructions. Additionally, we can use the recipe generation process to evaluate task trees that utilizes limited knowledge to solve novel problems, which possibly include never-before-seen ingredient combinations. To make the evaluation more concrete, we can use Intersection over Union (IoU) method together with a user study.


\section*{Acknowledgement}
\noindent This material is based upon work supported by the National Science Foundation under Grant Nos. 1910040 and 1812933.

\bibliographystyle{unsrt}
\bibliography{ref}

\end{document}